\documentclass[10pt, a4paper]{article}
\usepackage{authblk}
\usepackage{lrec-coling2024} 
\usepackage{graphicx}
\usepackage{multirow}
\PassOptionsToPackage{hyphens}{url}\usepackage{hyperref}
\usepackage{commath}

\title{A Neuro-Symbolic Approach to Monitoring Salt Content in Food}

\name{Anuja Tayal$^{1}$, Barbara Di Eugenio$^{1}$, Devika Salunke$^{2}$, Andrew D. Boyd$^{2}$, \\ 
{\bf  \large Carolyn A Dickens$^{3}$, Eulalia P Abril$^{4}$, Olga Garcia-Bedoya$^{3}$, Paula G Allen-Meares$^{3}$ }} 

\address{$^{1}$ Department of Computer Science 
$^{2}$ Department of Biomedical and Health Information Sciences \\
$^{3}$ Department of Medicine
$^{4}$ Department of Communications \\ 
University of Illinois Chicago, IL, USA 
\\ \{atayal4, bdieugen, dsalun2, boyda, cdickens,eulalia, ogarciab, pameares \}@uic.edu}

\abstract{
We propose a dialogue system that enables heart failure patients to inquire about salt content in foods and help them monitor and reduce salt intake. Addressing the lack of specific datasets for food-based salt content inquiries, we develop a template-based conversational dataset. The dataset is structured to ask clarification questions to identify food items and their salt content. Our findings indicate that while fine-tuning transformer-based models on the dataset yields limited performance, the integration of Neuro-Symbolic Rules significantly enhances the system’s performance. Our experiments show that by integrating neuro-symbolic rules, our system achieves an improvement in joint goal accuracy of over 20\% across different data sizes compared to naively fine-tuning transformer-based models.
\\ \newline }

\begin{document}

\maketitleabstract

\section{Introduction}
The excessive consumption of salt poses significant public health risks, contributing to diseases such as high blood pressure and heart failure \cite{salt-hf}. Reducing salt intake has been shown to mitigate these health issues.  In 2017, excessive sodium intake was associated with around three million deaths and a significant loss of healthy life years \cite{roth2018global}. Research, including clinical trials and population studies, supports the reduction of salt intake as a means to manage and prevent these conditions. Despite the clear benefits of sodium reduction, public understanding and action are lacking; only 58\% of individuals can accurately read sodium content on nutrition labels, and merely 44\% can classify food products as high or low in sodium based on standard labeling \cite{foodlabel}. This gap in knowledge and practice underscores the challenge of addressing dietary sodium intake, with only a handful of countries implementing effective public health interventions. 

Therefore, we aim to develop a dialogue system that enables patients to inquire about the salt content in various foods. This system especially aims to support heart failure patients, who must meticulously monitor and reduce their salt intake. More specifically,  African American individuals who are more prone to heart failure \cite{hfaa}, have a higher sensitivity to salt and face challenges like food deserts and higher consumption of junk foods. This necessitates a specialized dietary management approach to help them monitor and reduce their salt intake effectively. Furthermore, in \cite{itika-heart}, the authors show that African American patients with heart failure often focus on discussions related to salt and food during heart failure educational sessions, indicating a significant interest and need for information in this area. By providing a tool that facilitates easy access to information about salt content in foods, we aim to empower patients to make healthier dietary choices, thereby addressing a critical aspect of managing heart failure. Having said that, the dialog system can be used by anyone who wants to inquire about the salt content in food.

    \begin{figure}[t]
    \scriptsize
    \includegraphics[width=\linewidth]{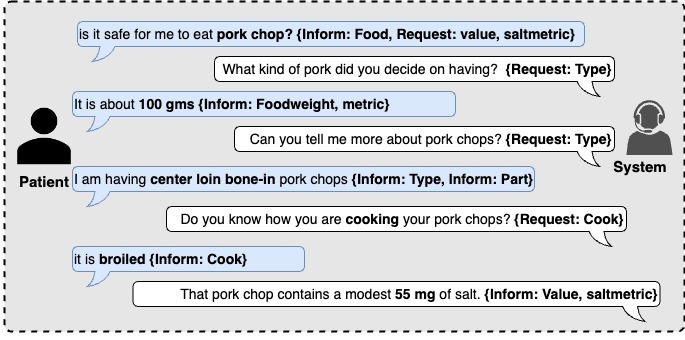}
    \caption{Sample Template Conversation which begins with the user asking about the salt content in food. The system asks clarification questions to determine the precise food item and its salt value.}
    \label{fig:sample-dialog}
    \end{figure}

Creating a dialog agent specialized in food-related conversations and nutrient information is challenging. This is primarily due to the lack of a conversational dataset specifically designed for this domain. Moreover, annotating the dataset is very costly and resource-intensive. To address this issue, we create a template-based conversational dataset (an example is shown in Figure \ref{fig:sample-dialog}) to identify various food items and their salt content. Our approach involves utilizing the USFDC (U.S. Food Data Central) \cite{abc} dataset, which provides detailed food descriptions along with their nutrient values. To enhance the system's ability to recognize the different components detailed in the food descriptions, we developed a food ontology. This ontology is constructed using the FoodOn \cite{foodon} framework and GloVe \cite{glove} embeddings, facilitating the identification of various attributes related to the food's cooking and preparation methods. Leveraging this ontology, we create a structured food conversational dataset (Figure \ref{fig:sample-dialog}). As it is a template-based dataset, annotating it is easy and effective. We model the dataset after the state-of-the-art task-oriented dialog dataset MultiWOZ \cite{multiwoz}.

To train the dialog system, we use the end-to-end dialog system PPTOD (Plug-and-Play Task-Oriented Dialogue System) Model \cite{pptod}. PPTOD extends the T5 framework, especially designed for task-oriented dialogue (TOD) applications. To accurately provide salt content information, our dialog system will address vague user queries, (such as "What is the salt content in pork chops?"), by asking clarification questions. This approach ensures the model understands the specific preparation and consumption context of the food item, allowing us to determine the precise salt content based on how the food is prepared (For ex - beef can be consumed raw, cooked, or as part of a dish like a burger). The dataset along with the trained model is publicly available.\footnote{\url{https://github.com/anujatayal/NS-Monitoring-Salt-Content-in-Food}}


Despite the remarkable proficiency of large pre-trained language models (PLMs) like GPT-3 and T5 \cite{few-shot,t5} in complex arithmetic reasoning tasks, they occasionally make calculation errors, especially as the mathematical operations in equations become more complex \cite{cot-reasoning}. Our observations align with these findings, notably that even after fine-tuning, the PPTOD model struggled to compute the salt values for various food items. Moreover, the USFDC provides salt content for standard food measurements, and users may not frame their queries in these standard terms. To rectify this, we propose \textbf{NS-PPTOD}, where we integrate  PPTOD model with neuro-symbolic rules. These rules are designed to harness the strengths of PLMs while compensating for their weaknesses, enabling the system to retrieve and accurately calculate the salt content from the database. This ensures the system's adaptability in responding to queries about salt content in both standard and non-standard food quantities, thereby offering accurate salt content information and enhancing the system’s overall accessibility and effectiveness.


Our experiments show that just fine-tuning a transformer model to predict salt content isn't enough. The integration of neuro-symbolic rules significantly enhances the system's performance, evidenced by a 20\% improvement in joint goal accuracy across different dataset sizes. This proves that combining pre-trained language models with neuro-symbolic rules is essential for better accuracy.
 
In summary, our contributions are

\begin{itemize}
    \item We propose to develop a food conversation dataset that includes clarifying questions to infer the correct food item and its salt content. 
    \item We finetuned PPTOD on our food conversation dataset using a few-shot approach.
    \item We propose NS-PPTOD which integrates PPTOD with Neuro-Symbolic rules to infer correct salt values across different food weights.
    \item We show a 20\% increase in joint accuracy compared to the finetuned PPTOD.
\end{itemize}

\section{Related Work}
\begin{itemize}
\item \textbf{HealthCare Dialog Systems}  Task-oriented dialogue systems have seen a significant rise in the healthcare sector, where they play a vital role in enhancing various aspects of healthcare. These systems are developed for a wide array of diseases including heart failure \cite{dil,itika-heart}, cancer \cite{cancer-ds}, mental disorders \cite{autism-ds}, public anxiety \cite{public-speaking-anxiety} etc. Their applications extend to several areas, including disease diagnosis \cite{automatic-diagnosis-pipeline}, patient education \cite{panini, itika-heart}, and health coaching \cite{yue-health} among others. A comprehensive survey of NLP literature conducted in \cite{healthcare-dialog-survey} provides an in-depth analysis of these diverse healthcare-oriented dialogue systems, examining them from a computational perspective and highlighting their varied end-users. 

\cite{itika-heart,hfchat} discuss the development of a dialog agent for self-care needs of heart failure patients, drawing upon insights from educational sessions. The work in \cite{woz-chi-persona-health-counceling} explores the Wizard of Oz (WOZ) technique to craft a persona-based health counseling dialog dataset. Additionally, recent advancements have seen the application of Large Language Models (LLMs) in responding to patient inquiries \cite{llm-medical-evaluation}, though the importance of safety is emphasized. Addressing the limitations in the medical knowledge of LLMs, the study in \cite{chatdoctor} undertakes the task of enhancing and fine-tuning the LLaMa model with a dataset of approximately 100,000 patient-doctor dialogues.

 \item \textbf{Pretrained Language Models (PLMs)} With the advancement in pre-trained language models (PLMs), different systems based on PLMs have been proposed including dialog systems \cite{sequicity,soloist}. PLMs excel in various tasks, approaching human-like performance. Yet, they struggle in mathematical reasoning, as noted in \cite{cot-reasoning}. \cite{lm-arithmetic} shows the limitations of LLMs with complex or lengthy numerical operations. For instance, GPT-3 \cite{few-shot} performs well in simple two-digit additions but falters with longer numbers. Similarly, even a fine-tuned T5 model struggles with the accurate addition or subtraction of lengthy numbers \cite{mathsinvestigating}, and the challenge escalates with numbers not covered in their training data.

\item \textbf{NeuroSymbolic AI} Integrating neuro-symbolic approaches offers a solution by combining the inference capabilities of symbolic systems with the robustness of neural networks, creating a composite AI framework adept at reasoning, learning, and cognitive modeling \cite{neurosymbolic}. This blend addresses the inherent weaknesses of each system, promising enhanced performance and robustness.

To address the generalization issues in neural networks, particularly in task-oriented dialogue systems, various neuro-symbolic methodologies have been investigated. \cite{schema} proposes  schema graphs to generalize across various unseen domains and tasks. In \cite{symbolic-injection}, the authors fine-tuned GPT-2 to generate the text and symbolic representations. DILOG \cite{dilog} employed inductive reasoning to formulate logical rules, enabling dialog policy training with minimal data to facilitate zero-shot domain transfer. \cite{conversational-commonsense}  used a neuro-symbolic approach to extract multi-hop reasoning and integrate commonsense in a dialog system. These strategies underscore the potential of neuro-symbolic integration to significantly improve the adaptability and efficacy of language models in complex and dynamic tasks. 

\item \textbf{Representing Food in Dialog Systems} Addressing the intricacies of food representation, FoodKG \cite{foodkg,personalised-food-recommendation} explored knowledge graphs to represent food. FoodKG \cite{foodkg} integrates information from diverse recipe collections and the US Food Data Central \cite{abc}, primarily focusing on template-based queries related to ingredients and recipes. However, this framework has limitations, notably in identifying only the primary item in food descriptions and missing key details like the type of food, cooking methods, or quantity, which are included in our methodology. \cite{mental-health} explored the role of recommending food to improve mental health. RecipeQA \cite{RecipeQA} explored multimodal question answering within the context of recipes, while CookDial \cite{cook2dial} provides a platform for users to navigate and query cooking recipes more effectively.

\end{itemize}
\section{Dataset Creation}
Given the absence of a specialized dataset for conversational inquiries about salt content and the challenges in dataset collection and annotation, in this section, we show in detail how we created the dataset. Dataset creation involves developing a template-based conversational framework to accurately identify food items and their salt content. First, we used the USFDC dataset \cite{abc}, and created an ontology using FoodOn \cite{foodon} and Glove \cite{glove} to describe the different components in the food description. By doing so, we were able to distinguish between different slot values. Using the ontology, we created a template-based conversational dataset that mimics human conversation, alternating between the user and the system turns.  We define an average of 7 slots while creating the dataset namely- food, cook, type, animal, part, foodweight, metric. 


\paragraph{Data Source } To construct the dataset, we leverage the extensive food descriptions and nutritional data from the USFDC database \cite{abc}. It is renowned for its broad representation of diverse food items and is publicly available. The dataset was created with careful consideration of cultural differences, sourcing its data from the U.S. Department of Agriculture (USDA).

Each food description in the USFDC database consist of unstructured, comma-separated text detailing ingredients, cooking methods, and cutting styles. This format lacks clarity on the significance of each component as demonstrated in Table \ref{table:food-descriptions} of Appendix \ref{sec:conversations}. To address this, we concatenated these segments using underscores, transforming each into a distinct entity to enhance data clarity and interpretation.

Furthermore, we faced the difficulty of distinguishing whether a food item is a primary ingredient or as part of a larger dish. Items like lettuce and cheese, for example, can be both independent food items and components in recipes like pizzas or burgers. To overcome this challenge, we developed a food ontology. This ontology aids in categorizing each food item more accurately, thus improving the overall understanding of the dataset.

\paragraph{Ontology Construction } To develop the food ontology, an initial framework is established using FoodON \cite{foodon}, focusing on key relations of \textit{food}, \textit{cook}, \textit{animal} and \textit{part}. These relations were chosen based on their significant impact on altering the salt content in various foods. Moving forward, these specific relations will be utilized to aid in creating and annotating the conversational dataset and pose clarifying questions to users to infer the salt amount in food. To address the limitations in the comprehensiveness of this initial ontology, pre-trained GloVe vectors \cite{glove} are utilized to identify words similar to those in the ontology, thus expanding its scope. However, this method inadvertently introduces some items unrelated to food, necessitating manual preprocessing to eliminate irrelevant elements and maintain a focus on food context.

Further refinement of the ontology was needed to incorporate the items that do not associate with existing relations. A new relation, \textit{type}, was created to integrate these components (For example type of cuisine, meat, other food ingredients etc). Drawing inspiration from FoodKG \cite{foodkg}, the first item in each food description is categorized under the \textit{food} relation. This enhanced ontology becomes a valuable tool for mapping each component of the comma-separated food descriptions to the relevant keys. In instances where multiple segments pertain to the \textit{type} relation, their values are concatenated to ensure consistency and clarity.

\subsection{Conversational Dataset Creation}
\label{dataset-creation}
\begin{table}[t]
\scriptsize
\begin{tabular}{p{.04\textwidth}|p{.4\textwidth}}
    \hline
key& Questions\\
   \hline
   \multirow{6}{*}{food}&What is the \{nutrient\} content in \{food\} ? \\
      &How much \{nutrient\} in \{food\}? \\
   & What is the \{nutrient\} content in \{cook\} \{food\}?\\
   & How much \{nutrient\} in \{foodWeight\} \{metric\} of \{food\}?\\
   & Can my partner with heart issues eat \{food\}?\\
 & Is \{food\} okay for heart patients? \\
 \hline
 \end{tabular}
    \caption{Sample Template Questions that user asks to begin the conversation}
    \label{tab:template}
\end{table}


Using the ontology and the food descriptions, we aim to develop a template-based conversational dataset that mimics human conversation. The conversation initiates with a user query about the salt content in a specific food item and alternates between the user and the system. The system poses clarification questions, drawing from ontology relations such as the type of food, cooking method, and portion size, which are crucial determinants of salt content.

Leveraging the task-oriented dialogue framework, each turn t is annotated with a belief state $B_t$, encompassing a list of slot-value pairs and action states \textit{inform} and \textit{request}. Figure \ref{fig:sample-dialog} illustrates a sample conversation highlighting the belief state and action state for each turn t. The figure demonstrates that within a single turn, it is possible to fill multiple slot values (the 3rd turn of the user). The dataset's format and annotations draw inspiration from the advanced task-oriented dialog dataset, MultiWOZ \cite{multiwoz}.   Employing MultiWOZ as a benchmark not only validates the dataset but also enhances its replicability for crafting conversational datasets for other nutrients. Using template-based approach streamlines the annotation process, ensuring uniformity and efficiency, and reducing the cost and time needed for external annotators.

\begin{table}[ht]
\centering
    \begin{tabular}{c|c}
    \hline
           &  \textbf{Dialogue Statistics} \\
         \hline
         \# Dialogues &87,425 \\
         \# Total  turns &525,392 \\
         Avg turns per dialogue &6 \\
         Avg slots &7 \\
         \hline
    \end{tabular}
    \caption{Dialogue Statistics of the template based conversation data}
    \label{table:stats}
\end{table}
We consider 3 types of turns in a conversation. We have tried to encapsulate the range of dynamics that can occur in dialogues, ensuring the system is robust enough to handle the fluidity of human conversation.

\begin{itemize}
    \item \textbf{Matching Answers}- This type involves turns where the user’s reply is directly pertinent to the system's question regarding a particular slot value. An example is when the system queries about the cooking method, and the user responds specifically about the food's cooking method (the 4th turn of the user in Figure \ref{fig:sample-dialog}).
    
    \item \textbf{Random Answers}- Occasionally, a user's reply may not correspond to the query posed by the system. For instance, if the system asks about the \textit{food type} and the user responds with information about the \textit{weight} of the item, as demonstrated in the user's second turn in Figure \ref{fig:sample-dialog}, the system needs to adapt. In such situations, the system should recognize and not repeat a question about the weight, since that information has already been provided. Instead, it should proceed to ask another question, possibly continuing to seek details about the \textit{food type}. In .45\% of conversations, a turn consists of a random answer.
    
    \item \textbf{Changing Answers}- People are very indecisive and often tend to change their responses. This category captures the scenarios where users might revise their previous responses. For example, if a user initially mentions that the cooking method is "pan-fried" but later changes it to "boiled," the system needs to update its understanding to reflect this new information, shifting its belief state from “pan-fried” to “boiled.” In .45\% of conversations, one user turn involves changing the answer.
\end{itemize}

The dataset is created by generating random conversations, in which, at each turn, the system's questions and the user's responses are randomly selected from the templates. The conversation starts with the user asking about the salt content in food. This initial question is informed by a user study of HFChat \cite{hfchat},  where participants frequently asked 3 categories of questions: 1) how much salt in \{food\} 2) Can I eat \{food\}? and 3) what kind of \{food\} can I eat?. These question types, along with similar ones, constitute the initial question, as exemplified in Table \ref{tab:template}. The system's objective is to  engage in the dialogue by asking clarification questions to ascertain the values of different slots (cook, type, weight etc). To keep the dialogue dynamic and realistic, questions related to these slots are presented randomly. The number of questions is limited; for instance, if the food in question is eaten raw, queries about cooking methods are omitted. Users might not be aware of all system-initiated questions, in which case default values for each slot are assumed. This led to the creation of approximately 87k template-based conversations, each comprising 3-4 exchanges between the user and the system. The statistics of this extensive dataset are detailed in Table~\ref{table:stats}. As the dataset size is very large, around 87k conversations, a few-shot method is used to train the model.

\section{Methodology}

Once the conversational dataset was created, we built the NS-PPTOD model  by fine-tuning PPTOD on the dataset using few-shot and integrating neuro-symbolic rules. 

\subsection {Plug and Play Task Oriented Dialog System (PPTOD)}
Leveraging the T5's model success, we adopted the PPTOD model for developing the task-oriented dialogue system. PPTOD extends the T5 framework, especially designed for task-oriented dialogue (TOD) applications, and pre-trained on a diverse range of dialog datasets spanning eleven different domains. PPTOD has integrated different TOD modules — Dialogue State Tracking (DST), Natural Language Understanding (NLU),  Dialogue Policy (POL) and Natural Language Generation (NLG) —into a single end-to-end architecture using a pipeline approach.

\begin{figure}[t]
    \scriptsize
    \includegraphics[width=\linewidth]{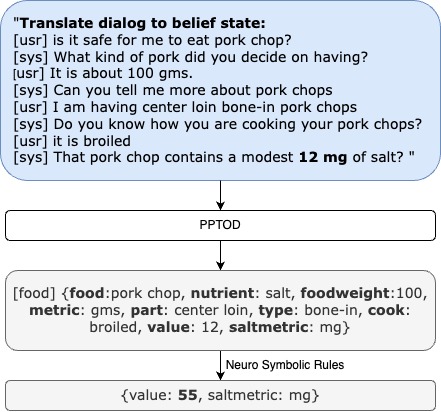}
    \caption{NS-PPTOD Model with Example: This example illustrates that PPTOD alone is not able to identify correct salt value for the food item}
    \label{fig:NSPPTOD}
\end{figure}

The PPTOD model is adept at in-context learning, employing customized prompts for each dialogue component, thus enhancing the relevance of model inputs to specific dialogue scenarios. Each training sample in PPTOD is represented as  ($x_t$, $y$, $z$), where $t$ $\in$ {NLU, DST, POL, NLG} specifies the type of Task-Oriented Dialogue (TOD) task to which the sample belongs. The term $x_t$ refers to the task-specific prompt, formatted as \textit{translate dialogue to A}:, where A represents different aspects such as \textit{user intent}, \textit{belief state}, \textit{dialogue act}, and \textit{system response} corresponding to the NLU, DST, POL, and NLG tasks respectively. The input dialogue context, a concatenation of all preceding dialogue utterances, encompassing both the system’s and the user’s contributions is represented by $y$ whereas $z$ indicates the target output text. This is exemplified in Figure \ref{fig:NSPPTOD}.
PPTOD is trained with a maximum likelihood objective and the loss function as shown in Equation \ref{loss-function} where $\theta$ is the model parameters. 

\begin{equation}
L_\theta = - \sum_{i=1}^{\abs{z}} \log P_\theta(z_i|z_{<i};x_t,y)
\label{loss-function}
\end{equation}

PPTOD also has an associated tokenizer, which supports a few-shot learning framework, enabling the system to identify new food-related terms not seen in training. We have fine-tuned it on only 1000 food-related dialogs using the same learning objective as PPTOD. 

We chose PPTOD  model  for its adaptability, and its ability to support  modular decomposition which in turn  facilitates  the  incorporation of specific rules. To further enhance the model, we incorporated neuro-symbolic techniques into the DST framework, aiming to boost the system’s interpretative capabilities and its adaptability to intricate dialogue contexts. This integration seeks to fortify the dialogue system, ensuring it remains robust and flexible in managing diverse conversations.

In summary, we chose PPTOD because:
\begin{itemize}
    \item PPTOD is a state-of-the-art (SOTA) model designed specifically for task-oriented dialogue and is based on the T5 model.
    \item PPTOD consists of a tokenizer making it possible to use few-shot approach to fine-tune it.
    \item PPTOD can be modularised to implement neuro-symbolic AI.
\end{itemize}

\subsection{NS-PPTOD}

NS-PPTOD is an integration of PPTOD with neuro-symbolic rules. The conversational dataset we created as described in section~\ref{dataset-creation} is used to finetune PPTOD. Given the extensive size of the dataset of 87k template-based conversations, a few-shot learning approach is used. Instead of utilizing the entire dataset, limited subsets of samples are selected randomly to finetune PPTOD. As we use a few-shot approach, T5-small is used as the base model to train PPTOD. PPTOD model is finetuned for 10 epochs, employing a batch size of 16 and varying the total dataset size across 100, 300, 500, and 1000 samples. Within the dataset, 10\% of the data was allocated as a development set, and another 10\% served as the test set. 

\begin{table}[t]
\scriptsize
    \begin{tabular}{{p{.08\textwidth} p{.05\textwidth} p{.05\textwidth} p{.05\textwidth} p{.05\textwidth} p{.05\textwidth}}}
    \hline
           &  \textbf{Train Size} &  \textbf{Epochs} & \textbf{Inform} & \textbf{Success} & \textbf{BLEU} \\
         \hline

         PPTOD &100&8&71.43&0&24.99 \\
         NS-PPTOD &100&-&88.90&77.80&22.50\\
         \hline
         PPTOD &300&7&75.00&5.00&34.30 \\
         NS-PPTOD &300&-&81.50&63.00&26.90\\
         \hline
         PPTOD &500&9&82.86&2.86&29.81 \\
         NS-PPTOD &500&-&74.50&58.10&28.90\\
         \hline
         PPTOD &1000&7&93.50&2.70&29.00 \\
         NS-PPTOD &1000&-&85.90&71.70&30.00\\
         
    \hline
    \end{tabular}
    \caption{Increase in performance when using NS-PPTOD compared to PPTOD.}
    \label{table:full-model}
\end{table}

During  the experiments - to be discussed  in Section~\ref{evaluation}, we observed that the model correctly identified slot values but struggled to determine the correct salt values. Additionally, the dataset primarily comprised salt values for standard food weights, like 100 grams, 3 ounces, 1 packet, etc., and lacked data for non-standard food weights that users might inquire about. PPTOD model is fine-tuned so that it accurately learns other slot values, even if it also learns some incorrect salt values. Subsequently, we employ a neuro-symbolic approach that involves two key methodologies to correct the salt values:
\begin{itemize}
\item \textbf{Retrieval of the accurate salt value from the database:} This step is crucial for standard food weights where exact values are available and can be directly obtained. (as shown in Figure \ref{fig:NSPPTOD})
\item \textbf{Mathematical calculation of the correct salt value for varying food weights:} This method is particularly beneficial for non-standard food weights, enabling the model to compute salt values based on weight. (In Figure \ref{fig:NSPPTOD}, if the user requests the salt value for a different food weight, instead of the standard 100 grams of pork chops.)
\end{itemize}

Specifically, upon determining the slot values, the system queries the database for the salt content. If the database contains the salt value, it is then retrieved. In cases where the salt value is not available in the database, it is calculated mathematically, based on the weight of the food. We demonstrate this in Figure \ref{fig:NSPPTOD}. When the dialog context and prompt, labeled \textit{Translate dialog to belief state}, are processed through PPTOD's DST model to determine belief states, the model successfully infers all slot values except for the salt value (12). The correct salt value (81) is then retrieved from the database by applying Neuro-Symbolic rules.

\begin{table}[t]
\centering
    \begin{tabular}{|p{.1\textwidth} p{.07\textwidth} p{.07\textwidth} | p{.13\textwidth}|}
    \hline
           \textbf{Train Size} &  \textbf{Epochs} & \multicolumn{2}{c|} {\textbf{Joint Accuracy}} \\
           & & \textbf{PPTOD} & \textbf{NS-PPTOD} \\
         \hline
         100&6&55.56&73.08 \\
         300&4&51.92&72.8 \\
         500&6&58.75&83.2 \\
         1000&6&58.53 &85.2 \\
    \hline
    \end{tabular}
    \caption{Increase in Joint Accuracy when using NS-PPTOD compared to PPTOD across different training sizes}
    \label{table:joint-accuracy}
\end{table}  
\section{Evaluation}
\label{evaluation}
The evaluation of NS-PPTOD encompasses two task-oriented tasks. The first task involves end-to-end dialog modeling, assessed using metrics such as inform rate, success rate, and BLEU score \cite{bleu}. The second task, the Dialog State Tracking (DST) module of PPTOD is evaluated through joint-accuracy. This evaluation was conducted over sample sizes of 100, 300, 500, and 1000.


\paragraph{End-to-End Dialog Modeling} After seven epochs of training, the PPTOD model demonstrated a high inform accuracy rate of 93.5\% across 1000 samples, as detailed in Table \ref{table:full-model}. Inform rate reflects the model's adeptness in identifying slot values and the target goal slot. However, its success rate in accurately determining correct salt values was notably low, standing at just 2.7\%, a point further highlighted in Table \ref{table:full-model}. This pattern of low success rate was consistent across other training sizes of 100, 300, and 500 samples. The limited 2\% success rate is attributed to the PPTOD model's tendency to predict values at random.

Implementing the NS-PPTOD model resulted in a substantial enhancement, achieving success rate of 71.7\%. This improvement was not just limited to the training size of 1000 samples but was also observed consistently across the smaller training sizes. The integration of neuro-symbolic rules with PPTOD evidently plays a crucial role in enhancing the model's capability to accurately predict and determine the correct salt values.

\paragraph{DST Module} The evaluation also included the Dialog State Tracking (DST) module of PPTOD, trained for recognizing different belief states such as food, cook, type, weight, and value. Joint accuracy of 58.53\% was achieved when PPTOD was used. This performance notably increased to 85.2\% for 1000 samples with the addition of neuro-symbolic rules. Similar improvements in joint accuracy were observed for other training sizes. Table \ref{table:joint-accuracy} displays the enhanced joint-accuracy achieved by using NS-PPTOD compared to PPTOD.

\paragraph{Analysis} The improvement in both success rate and joint accuracy can be attributed to a key difference in approach. PPTOD, on its own, tends to memorize values instead of effectively retrieving them from the database, a critical process for accurately determining salt content. However, the application of neuro-symbolic rules in conjunction with PPTOD enhances its capability, enabling it to effectively retrieve values from the database. 

\subsection{Comparison with ChatGPT}

With the advent of ChatGPT, questions have arisen about the necessity of systems such as our NS-PPTOD. To address this,we conducted  a comparative analysis between NS-PPTOD and ChatGPT, to highlight the distinct capabilities and applications of NS-PPTOD that are not achievable by ChatGPT.

In this comparison, we specifically  focused on their responses to queries about the salt content in foods. Appendix \ref{sec:chatgpt} illustrates ChatGPT's response to the prompt \textit{What is the salt amount in a pork chop?}. ChatGPT's responses are generally comprehensive, explaining variations in salt quantity due to different cooking methods and weights, and often include an average value. NS-PPTOD, conversely, poses targeted clarification questions to precisely identify both the food item and its salt content.

There is also a marked difference in the readability of responses from these two systems. Readability assessments, SMOG \cite{smog}, Flesch-Kincaid Grade Level (FKGL) \cite{fkre} , and Flesch-Kincaid Reading Ease (FKRE) \cite{fkre}, which evaluates text comprehensibility in educational and informational contexts, were used for this comparison. The SMOG and FKGL estimates the education level needed to understand a text, while the FKRE measures the ease of comprehension.

As indicated in Table \ref{table:readability}, understanding ChatGPT's content requires an 11th-grade reading level, whereas NS-PPTOD's content is accessible to those at a 2nd-grade reading level. NS-PPTOD's responses align with recommendations for content to be understandable at a 5th-grade reading level or lower \cite{reading-level}. This approach is particularly suitable for the intended  audience of our system, the patients served by the hospital at our institution who are mostly of lower socio-economic status (and mostly minorities); this ensures the information is accessible and understandable to a wider audience.


\begin{table}[ht]
\centering
    \begin{tabular}{|c|c|c|c|}
    \hline
           &  \textbf{SMOG} & \textbf{FKGL} & \textbf{FKRE} \\
         \hline 
         ChatGPT & 11.8 & 11.1 & 51.28 \\
         \hline
         NS-PPTOD &0 &2.5&95.17 \\
    \hline
    \end{tabular}
    \caption{Readability Comparison of NS-PPTOD with ChatGPT: A higher FKRE score denotes better readability. Lower scores in SMOG and FKGL indicate better text readability.}
    \label{table:readability}
\end{table}  

\section{Conclusion and Future Work}
This work introduces NS-PPTOD, a conversational dialogue system specifically designed to help patients, particularly African-American individuals suffering from heart failure, in accurately identifying the salt content in various foods. Managing salt intake is crucial for these patients, and NS-PPTOD offers a specialized solution to aid in their dietary management. The study reveals that simply applying fine-tuning techniques to a transformer model for predicting salt content is inadequate for achieving high accuracy. However, the integration of neuro-symbolic rules into the system resulted in a significant performance boost, enhancing accuracy by 20\% across different dataset sizes. This enhancement underscores the necessity of combining pretrained language models with neuro-symbolic rules to achieve superior accuracy.

Looking ahead, there are plans to conduct an in-depth study comparing the performance of NS-PPTOD with a system based on ChatGPT. This study will involve African American patients with heart failure interacting with both systems in a within-subject setting. Utilizing pre and post-interaction surveys, the aim is to conduct both qualitative and quantitative analyses. These analyses will evaluate and contrast the effectiveness of each system offering insights into their respective impact in real-world scenarios.

\section{Acknowledgement}
For financial support,  we thank the DPI Cycle 1 Seed Funding Program Award and NSF award IIS 2232307

\bibliographystyle{lrec-coling2024-natbib}
\bibliography{lrec-coling2024-example}

\bibliographystylelanguageresource{lrec-coling2024-natbib}
\bibliographylanguageresource{languageresource}
\appendix

\section{Sample Food Descriptions }
\label{sec:conversations}
\begin{table}[h]
\scriptsize
    \begin{tabular}{p{.35\textwidth}|p{.05\textwidth}}
    \hline
          \textbf{Food Descriptions} &  \textbf{Salt Value} \\
         \hline

         Pork, fresh, loin, top loin (chops), boneless, separable lean and fat, raw &48\\
         Pork, fresh, loin, center loin (chops), bone-in, separable lean and fat, cooked, broiled &55\\
         Pork, fresh, blade, (chops), boneless, separable lean and fat, cooked, broiled &58\\
         Pork, fresh, loin, sirloin (chops or roasts), boneless, separable lean only, raw &63 \\
         Pork, fresh, loin, blade (chops), bone-in, separable lean only, cooked, broiled &76\\
         
    \hline
    \end{tabular}
    \caption{Unstructured Food Descriptions along with their salt value for 100gms of food}
    \label{table:food-descriptions}
\end{table}  

\section{Sample ChatGPT Response}
\label{sec:chatgpt}
ChatGPT response to the prompt \textbf{How much salt in pork chops?} 

\textit{The salt content in pork can vary significantly based on how it is prepared. Fresh, unprocessed pork naturally contains a small amount of sodium, roughly 60 to 70 milligrams of sodium per 3-ounce serving. However, processed pork products like bacon, ham, sausages, and deli meats can have a much higher sodium content due to added salt during processing for flavor and preservation. For instance, a slice of bacon can have between 150 to 450 milligrams of sodium, and a serving of ham can contain over 1,000 milligrams.}

\textit{For specific sodium content, it's best to check the nutrition labels on packaged pork products. For fresh pork, nutrition databases or guidelines from health organizations can provide general estimates. Remember, if you're watching your sodium intake, opting for fresh, unprocessed pork and limiting processed pork products can help manage your sodium consumption.}

\end{document}